\documentclass[letterpaper]{article} 
\usepackage{aaai23}  
\usepackage{times}  
\usepackage{helvet}  
\usepackage{courier}  
\usepackage[hyphens]{url}  
\usepackage{graphicx} 
\usepackage{natbib}  
\usepackage{caption} 
\usepackage{algorithm}
\usepackage{algorithmic}

\usepackage{amsmath,amsfonts,bm, amsthm, mathrsfs, mathtools}

\newtheorem{problem}{Problem}

\numberwithin{theorem}{section}
\numberwithin{lemma}{section}
\numberwithin{proposition}{section}
\numberwithin{corollary}{section}

\def\eqref#1{equation~\ref{#1}}

\def\1{\bm{1}}

\def\vg{{\bm{g}}}
\def\vh{{\bm{h}}}

\def\vu{{\bm{u}}}

\def\vw{{\bm{w}}}
\def\vx{{\bm{x}}}
\def\vy{{\bm{y}}}

\DeclareMathAlphabet{\mathsfit}{\encodingdefault}{\sfdefault}{m}{sl}
\SetMathAlphabet{\mathsfit}{bold}{\encodingdefault}{\sfdefault}{bx}{n}

\def\gD{{\mathcal{D}}}

\def\gH{{\mathcal{H}}}
\def\gI{{\mathcal{I}}}

\def\gL{{\mathcal{L}}}

\def\gW{{\mathcal{W}}}
\def\gX{{\mathcal{X}}}
\def\gY{{\mathcal{Y}}}
\def\gZ{{\mathcal{Z}}}

\def\sR{{\mathbb{R}}}

\newcommand{\E}{\mathbb{E}}

\DeclareMathOperator*{\argmin}{arg\,min}

\usepackage{xcolor}     
\usepackage{makecell}
\usepackage{booktabs}       
\usepackage{amssymb}
\usepackage{comment}
\usepackage{newfloat}
\usepackage{listings}
\DeclareCaptionStyle{ruled}{labelfont=normalfont,labelsep=colon,strut=off} 
\floatstyle{ruled}
\newfloat{listing}{tb}{lst}{}
\floatname{listing}{Listing}
\title{Post-hoc Uncertainty Learning using a Dirichlet Meta-Model}
\author{
    Maohao Shen\textsuperscript{\rm 1}, 
    Yuheng Bu\textsuperscript{\rm 2}, 
    Prasanna Sattigeri\textsuperscript{\rm 3},\\
    Soumya Ghosh\textsuperscript{\rm 3}, 
    Subhro Das\textsuperscript{\rm 3},
    Gregory Wornell\textsuperscript{\rm 1}
}
\affiliations{
    \textsuperscript{\rm 1} Massachusetts Institute of Technology\\
    \textsuperscript{\rm 2} University of Florida\\
    \textsuperscript{\rm 3} MIT-IBM Watson AI Lab, IBM Research\\

    maohao@mit.edu
}

\usepackage{bibentry}

\begin{document}

\maketitle

\begin{abstract}

It is known that neural networks have the problem of being over-confident when directly using the output label distribution to generate uncertainty measures. Existing methods mainly resolve this issue by retraining the entire model to impose the uncertainty quantification capability so that the learned model can achieve desired performance in accuracy and uncertainty prediction simultaneously. However, training the model from scratch is computationally expensive and may not be feasible in many situations. In this work, we consider a more practical post-hoc uncertainty learning setting, where a well-trained base model is given, and we focus on the uncertainty quantification task at the second stage of training. We propose a novel Bayesian meta-model to augment pre-trained models with better uncertainty quantification abilities, which is effective and computationally efficient. Our proposed method requires no additional training data and is flexible enough to quantify different uncertainties and easily adapt to different application settings, including out-of-domain data detection, misclassification detection, and trustworthy transfer learning. We demonstrate our proposed meta-model approach's flexibility and superior empirical performance on these applications over multiple representative image classification benchmarks.

\end{abstract}

\section{Introduction}\label{sec:intro}

Despite the promising performance of deep neural networks achieved in various practical tasks~\citep{simonyan2014very, ren2015faster, hinton2012deep, mikolov2010recurrent, alipanahi2015predicting, litjens2017survey}, uncertainty quantification (UQ) has attracted growing attention in recent years to fulfill the emerging demand for more robust and reliable machine learning models, as UQ aims to measure the reliability of the model's prediction quantitatively. Accurate uncertainty estimation is especially critical for the field that is highly sensitive to error prediction, such as autonomous driving~\citep{bojarski2016end} and medical diagnosis~\citep{begoli2019need}.

Most state-of-the-art approaches~\citep{gal2016dropout, lakshminarayanan2017simple, malinin2018predictive, van2020simple} focus on building a deep model equipped with uncertainty quantification capability so that a single deep model can achieve both desired prediction and UQ performance simultaneously. However, such an approach for UQ suffers from practical limitations because it either requires a specific model structure or explicitly training the entire model from scratch to impose the uncertainty quantification ability.  A more realistic scenario is to quantify the uncertainty of a pretrained model in a post-hoc manner due to practical constraints. For example, (1) compared with prediction accuracy and generalization performance, uncertainty quantification ability of deep learning models are usually considered with lower priority, especially for profit-oriented applications, such as recommendation systems; (2) some applications require the models to impose other constraints, such as fairness or privacy, which might sacrifice the UQ performance; (3) for some applications such as transfer learning, the pretrained models are usually available, and it might be a waste of resources to train a new model from scratch.  

Motivated by these practical concerns, we focus on tackling the post-hoc uncertainty learning problem, i.e., given a pretrained model, how to improve its UQ quality without affecting its predictive performance. 
Prior works on improving uncertainty quality in a post-hoc setting have been mainly targeted towards improving calibration \citep{guo2017calibration,kull2019beyond}. These approaches typically fail to augment the pre-trained model with the ability to capture different sources of uncertainty, such as epistemic uncertainty, which is crucial for applications such Out-of-Distribution (OOD) detection. Several recent works \citep{chen2019confidence,jain2021deup} have adopted the meta-modeling approach, where a meta-model is trained to predict whether or not the pretrained model is correct on the validation samples. These methods still rely on a point estimate of the meta-model parameters, which can be unreliable, especially when the validation set is small. 

In this paper, we propose a novel Bayesian meta-model-based uncertainty learning approach to mitigate the aforementioned issues. Our proposed method requires no additional data other than the training dataset and is flexible enough to quantify different kinds of uncertainties and easily adapt to different application settings.
Our empirical results provides crucial insights regarding meta-model training: (1) The diversity in feature representations across different layers is essential for uncertainty quantification, especially for out-of-domain (OOD) data detection tasks; (2) Leveraging the Dirichlet meta-model to capture different uncertainties, including total uncertainty and epistemic uncertainty; (3) There exists an over-fitting issue in uncertainty learning similar to supervised learning that needs to be addressed by a novel validation strategy to achieve better performance. Furthermore, we show that our proposed approach has the flexibility to adapt to various applications, including OOD detection, misclassification detection, and trustworthy transfer learning. 


\section{Related Work}

Uncertainty Quantification methods can be broadly classified as \textit{intrinsic} or \textit{extrinsic} depending on how the uncertainties are obtained from the machine learning models. Intrinsic methods encompass models that inherently provide an uncertainty estimate along with its predictions. Some intrinsic methods such as neural networks with homoscedastic/heteroscedastic noise models \citep{wakefield2013bayesian} and quantile regression \cite{koenker1978regression} can only capture \textit{Data} (aleatoric) uncertainty. Many applications including out-of-distribution detection, requires capturing both \textit{Data} (aleatoric) and \textit{Model} (epistemic) accurately. Bayesian methods such as Bayesian neural networks (BNNs) \cite{neal2012bayesian,blundell15,welling2011bayesian} and Gaussian processes \cite{gpbook} and ensemble methods~\citep{lakshminarayanan2017simple} are well known examples of intrinsic methods that can 
quantify both uncertainties. However, Bayesian methods and ensembles can be quite expensive and require several approximations to learn/optimize in practice~\citep{mackay1992practical,kristiadi2021learnable,welling2011bayesian}. Other approaches attempt to alleviate these issues by directly parameterizing a Dirichlet prior distribution over the categorical label proportions via a neural network~\citep{malinin2018predictive, sensoy2018evidential, malinin2019reverse, nandy2020towards, charpentier2020posterior, joo2020being}. 

Under model misspecification, Bayesian approaches are not well-calibrated and can produce severely miscalibrated uncertainties. In the particular case of BNNs, sparsity-promoting priors have been shown to produce better-calibrated uncertainties, especially in the small data regime \cite{ghosh2019model}, and somewhat alleviate the issue. Improved approximate inference methods and methods for prior elicitation in BNN models are active areas of research for forcing BNNs to produce better-calibrated uncertainties. Frequentist methods that approximate the jackknife have also been proposed to construct calibrated confidence intervals~\cite{alaa2019discriminative}.


For models without an inherent notion of uncertainty, extrinsic methods are employed to extract uncertainties in a post-hoc manner. The post-hoc UQ problem is still under-explored, and few works focus on tackling this problem. 
One such approach is to build auxiliary or meta-models, which have been used successfully to generate reliable confidence measures (in classification) \citep{chen2019confidence}, prediction intervals (in regression), and to predict performance metrics such as accuracy on unseen and unlabeled data~\citep{elder2020learning}. Similarly, DEUP~\citep{jain2021deup} trains an error predictor to estimate the epistemic uncertainty of the pretrained model in terms of the difference between generalization error and aleatoric uncertainty. LULA~\citep{kristiadi2021learnable} trains additional hidden units building on layers of MAP-trained model to improve its uncertainty calibration with Laplace approximation. However, many of these methods require additional data samples that are either validation or out of distribution dataset to train or tune the hyper-parameter, which is infeasible when these data are not available. Moreover, they are often not flexible enough to distinguish epistemic uncertainty and aleatoric uncertainty, which are known to be crucial in various learning applications~\citep{hullermeier2021aleatoric}. In contrast, our proposed method does not require additional training data or modifying the training procedure of the base model.

\section{Problem Formulation} \label{sec:problem}
We focus on classification problems in this paper. Let $\gZ = \gX \times \gY$, where $\gX$ denotes the input space, and $\gY=\{1,\cdots,K\}$ denotes the label space. Given a base-model training set $\mathcal{D}_B=\left\{\boldsymbol{x}^B_{i}, y^B_{i}\right\}_{i=1}^{N_B}\in \gZ^{N_B} $ containing i.i.d. samples generated from the distribution $P^{B}_{Z}$, a pretrained base model $\vh \circ \Phi:\gX \rightarrow \Delta^{K-1}$ is constructed, where $\Phi: \gX \rightarrow \sR^{l}$ and $\vh: \sR^{l} \rightarrow \Delta^{K-1}$ denote two complementary components of the neural network. More specifically,  $\Phi\left(\vx\right) = \Phi\left(\vx; \vw_{\phi}\right)$ stands for the intermediate feature representation of the base model, and the model output $\vh\left(\Phi(\vx)\right) = \vh\left(\Phi(\vx; \vw_{\phi}); \vw_h\right)$ denotes the  predicted label distribution $P_B\left( \vy \mid \Phi(\vx)\right) \in \Delta^{K-1}$ given input sample $\vx$, where $\left(\vw_{\phi}, \vw_h\right)\in \gW$ are the parameters of the pretrained base model.

The performance of the base model is evaluated by a non-negative loss function $\ell_B: \gW \times \gZ \to \sR_+$, e.g., cross entropy loss. Thus, a standard way to obtain the pretrained base model is by minimizing the empirical risk over $\mathcal{D}_B$, i.e., $\gL_B(\vh \circ \Phi,\gD_B) \triangleq \frac{1}{N_B}\sum_{i=1}^{N_B} \ell_B \left(\vh \circ \Phi, (\vx^B_i,y^B_i) \right)$.   


Although the well-trained deep base model is able to achieve good prediction accuracy, the output label distribution $P_B\left(\vy \mid \Phi(\vx)\right)$ is usually  unreliable for uncertainty quantification, i.e., it can be overconfident or poorly calibrated. Without retraining the model from scratch, we are interested in improving the uncertainty quantification performance in an efficient post-hoc manner. We utilize a meta-model $\vg : \sR^{l} \rightarrow \mathcal{\tilde{Y}}$ with parameter $\vw_{g}\in \gW_g$ building on top of the base model. The meta-model shares the same feature extractor from the base model and generates an output $\tilde{\vy} = \vg\left(\Phi(\vx); \vw_g\right)$, where $\tilde{\vy}\in \mathcal{\tilde{Y}}$ can take any form, e.g., a distribution over $\Delta^{K-1}$ or a scalar. Given a meta-model training set $\mathcal{D}_M=\left\{\boldsymbol{x}^M_{i}, y^M_{i}\right\}_{i=1}^{N_M}$ with i.i.d. samples from the distribution $P^{M}_{Z}$, our goal is to obtain the meta-model by optimizing a training objective $\gL_M(\vg \circ \Phi,\gD_M) \triangleq \frac{1}{N_M}\sum_{i=1}^{N_M} \ell_M \left(\vg \circ \Phi, (\vx^M_i,y^M_i) \right)$, where $\ell_M: \gW_g \times \gW_{\phi} \times \gZ \to \sR_+$ is the loss function for the meta-model. 


In the following, we formally introduce the post-hoc uncertainty learning problem using meta-model.
\begin{problem}\label{prob:general}[Post-hoc Uncertainty Learning by Meta-model] Given a base model $\vh \circ \Phi$ learned from the base-model training set $\gD_B$, the uncertainty learning problem by meta-model is to learn the function $\vg$ using the meta-model training set $\gD_M$ and the shared feature extractor $\Phi$, i.e.,
\begin{equation}
\vg^* = \argmin_{\vg} \gL_M(\vg \circ \Phi,\gD_M),
\end{equation}
such that the output from the meta-model $\tilde{\vy} = \vg\left(\Phi(x)\right)$ equipped with an uncertainty metric function $\vu : \mathcal{\tilde{Y}} \rightarrow \sR$ is able to generate a robust uncertainty score $\vu\left(\tilde{\vy}\right)$.
\end{problem}

Next, the most critical questions are how the meta-model should use the information extracted from the pretrained base model, what kinds of uncertainty the meta-model should aim to quantify, and finally, how to train the meta-model appropriately. 




\section{Method}\label{sec:method}

In this section, we specify the post-hoc uncertainty learning framework defined in Problem~\ref{prob:general}. First, we introduce the structure of the meta-model. Next, we discuss the meta-model training procedure, including the training objectives and a validation trick. Finally, we define metrics for uncertainty quantification used in different applications.

The design of our proposed meta-model method is based on three high-level insights: (1) Different intermediate layers of the base model usually capture various levels of feature representation, from low-level features to high-frequency features, e.g., for OOD detection task, the OOD data is unlikely to be similar to in-distribution data across all levels of feature representations. Therefore, it is crucial to leverage the diversity in feature representations to achieve better uncertainty quantification performance. (2) Bayesian method is known to be capable of modeling different types of uncertainty for various uncertainty quantification applications, i.e., total uncertainty and epistemic uncertainty. Thus, we propose a Bayesian meta-model to parameterize the Dirichlet distribution, used as the conjugate prior distribution over label distribution. (3) We believe that the overconfident issue of the base model is caused by over-fitting in supervised learning with cross-entropy loss. In the post-hoc training of the meta-model, a validation strategy is proposed to improve the performance of uncertain learning instead of prediction accuracy.

\begin{figure}[t]
  \centering
  \includegraphics[width=1\linewidth]{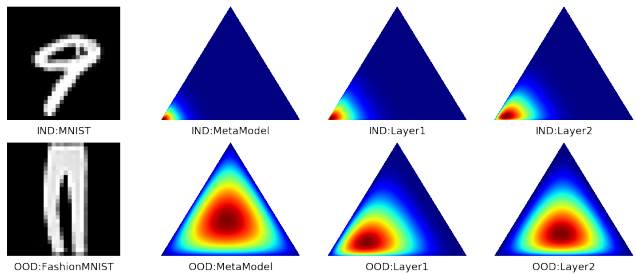}
  \vspace{-0.5em}
  \captionsetup{font=small}
  \caption{A toy example of our proposed meta-model method in OOD detection application shows the diversity of features in different layers. MetaModel utilizes two intermediate features, while Layer1 and Layer2 only are trained with one individual feature.}
  \label{fig: toy_problem}
\end{figure}

Before discussing the details of our proposed method, we want to use a toy example of the OOD detection task shown in Figure~\ref{fig: toy_problem} to elaborate our insights. The goal is to improve the OOD (FashionMNIST) detection performance of a LeNet base model trained on MNIST. The meta-model takes base-model intermediate feature representation as input and parameterizes a Dirichlet distribution over the probability simplex. We train three different meta-models using both intermediate layer features, using either of the two features, respectively, and then visualize and compare the output Dirichlet distribution on the simplex. Specifically, we take the three dominant classes with the largest logits to approximately visualize the Dirichlet distribution over the probability simplex. We observe that the meta-model outputs a much sharper distribution for the in-distribution sample than the OOD sample. Moreover, compared to the meta-model trained with one feature, the meta-model trained with multiple intermediate layers can further distinguish the two distributions on simplex. They generate sharper distribution for the in-distribution sample while exhibiting a more uniform distribution for the OOD sample, which strongly supports our key claims. 


\subsection{Meta-Model Structure}\label{subsec: model}
The proposed meta-model consists of multiple linear layers $\{\vg_j\}_{j=1}^{m}$ attached to different intermediate layers from the base model, and a final linear layer $\vg_{c}$ that combines all the features and generates a single output. Specifically, given an input sample $\vx$, denote the multiple intermediate feature representation extracted from the base-model as $\{\Phi_j\left(\vx\right)\}_{j=1}^{m}$. For each intermediate base-feature $\Phi_j$, the corresponding linear layer will construct a low-dimensional meta-feature $\{\vg_j\left(\Phi_j\left(\vx\right)\right)\}_{j=1}^{m}$. Then, the final linear layer of the meta-model takes the multiple meta-features as inputs and generates a single output, i.e., $\tilde{\vy} = \vg\left(\{\Phi_j\left(\vx\right)\}_{j=1}^{m}; \vw_g\right) = \vg_c\left(\{\vg_j\left(\Phi_j\left(\vx\right)\right)\}_{j=1}^{m}; \vw_{g_c}\right) $. In practice, the linear layers $\vg_{i}$ and $\vg_{c}$ only consist of fully connected layers and activation function, which ensures the meta-model has a much simpler structure and enables efficient training. 

\begin{figure}[t]
  \centering
  \includegraphics[width=1\linewidth]{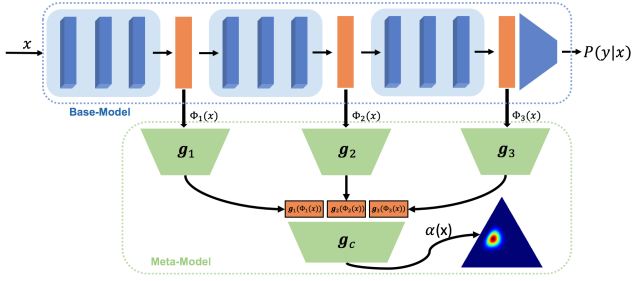}
  \vspace{-0.5em}
  \captionsetup{font=small}
  \caption{Meta-Model structure}
  \label{fig: structure}
\end{figure}

Given an input sample $\vx$, the base model outputs a conditional label distribution $P_B\left(\vy | \Phi(\vx)\right) \in \Delta^{K-1}$, corresponding to a single point in the probability simplex. However, such label distribution $P_B\left(\vy | \Phi(\vx)\right)$ is a point estimate, which only shows the model's uncertainty about different classes but cannot reflect the uncertainty due to the lack of knowledge of a given sample, i.e., the epistemic uncertainty. To this end, we adopt the Dirichlet technique commonly used in the recent literature~\citep{malinin2018predictive, malinin2019reverse, nandy2020towards, charpentier2020posterior} in order to better quantify the epistemic uncertainty. Let the label distribution as a random variable over probability simplex, denote as $\boldsymbol{\pi} = [\pi_1, \pi_2, ..., \pi_k]$, 
the Dirichlet prior distribution is the conjugate prior of the categorical distribution, i.e., 
\begin{equation}
\operatorname{Dir}(\boldsymbol{\pi} | \boldsymbol{\alpha}) \triangleq \frac{\Gamma\left(\alpha_{0}\right)}{\prod_{c=1}^{K} \Gamma\left(\alpha_{c}\right)} \prod_{c=1}^{K} \pi_{c}^{\alpha_{c}-1}, \alpha_c>0,\alpha_0 \triangleq \sum_{c=1}^K \alpha_c,
\end{equation}
Our meta-model $\vg$ explicitly parameterize the posterior Dirichlet distribution, i.e, 
\begin{equation*}
    q(\boldsymbol{\pi}|\Phi(\vx);\vw_g ) \triangleq \operatorname{Dir}(\boldsymbol{\pi} | \boldsymbol{\alpha}(\vx)), \  \boldsymbol{\alpha}(\vx) = e^{\vg\left(\Phi(\vx); \vw_g\right)},
\end{equation*}
where the output of our meta-model is $\tilde{\vy} = \log{\boldsymbol{\alpha}(\vx)}$, and $\boldsymbol{\alpha}(\vx) = [\alpha_1(\vx), \alpha_2(\vx), ..., \alpha_k(\vx)]$ is the concentration parameter of the Dirichlet distribution given an input $\vx$. The overall structure of the meta-model is shown in Figure~\ref{fig: structure}.

\subsection{Uncertainty Learning}
\paragraph{Training Objective}\label{subsec: objective}
From Bayesian perspective, the predicted label distribution using the Dirichlet meta-model is given by the expected categorical distribution: 
\begin{equation}
q(y=c|\Phi(\vx); \vw_g) = \E_{q(\boldsymbol{\pi}|\Phi(\vx); \vw_g)} [P(y=c|\boldsymbol{\pi})] = \frac{\alpha_c(\vx)}{\alpha_0(\vx)},
\end{equation}
where $\alpha_0=\sum_{c=1}^K \alpha_c$ is the precision of the Dirichlet distribution.

The true posterior of the categorical distribution over sample $(\vx,y)$ is $P(\boldsymbol{\pi}| \Phi(\vx), y) \propto P(y | \boldsymbol{\pi}, \Phi(\vx)) P(\boldsymbol{\pi} | \Phi(\vx))$, which is difficult to evaluate. Instead, we utilize a variational inference technique used in~\citep{joo2020being} to generate a variational distribution $q(\boldsymbol{\pi}| \Phi(\vx); \vw_g)$ parameterized by Dirichlet distribution with meta-model to approximate the true posterior distribution $P(\boldsymbol{\pi}| \Phi(\vx), y)$, then minimize the KL-divergence $\mathrm{KL}\left(q(\boldsymbol{\pi}| \Phi(\vx); \vw_g) \| P(\boldsymbol{\pi}| \Phi(\vx), y)\right)$, which is equivalent to maximize the evidence lower bound (ELBO) loss (derivation is provided in Appendix~\ref{sec:ELBO-appendix}), i.e.,
\begin{align}
\gL_{\mathrm{VI}}(\vw_g) &=\frac{1}{N_M}\sum_{i=1}^{N_M} \mathbb{E}_{q(\boldsymbol{\pi}| \Phi(\vx_i); \vw_g)}[\log P(y_i \mid \boldsymbol{\pi}, \boldsymbol{x}_i)] \nonumber \\
&\quad - \lambda\cdot\mathrm{KL}\left(q(\boldsymbol{\pi}| \Phi(\vx_i); \vw_g) \| P(\boldsymbol{\pi} | \Phi(\vx_i))\right)\\
&= \frac{1}{N_M}\sum_{i=1}^{N_M} \psi\left(\alpha^{(i)}_{y_i}\right)-\psi\left(\alpha^{(i)}_{0}\right)  \nonumber\\
&\quad - \lambda\cdot \mathrm{KL}\left( \operatorname{Dir}(\boldsymbol{\pi}| \boldsymbol{\alpha}^{(i)}) \| \operatorname{Dir}(\boldsymbol{\pi} | \boldsymbol{\beta})\right),
\end{align}
where $\boldsymbol{\alpha}^{(i)}$ is the Dirichlet concentration parameter parameterized by the meta-model, i.e., $\boldsymbol{\alpha}^{(i)} = e^{\left(\vg\left(\Phi(\vx_i); \vw_g\right)\right)}$, $\boldsymbol{\psi}$ is the digamma function, and $\boldsymbol{\beta}$ is the predefined concentration parameter of prior distribution. In practice, we simply let $\boldsymbol{\beta} = [1,\cdots,1]$. The likelihood term encourages the categorical distribution to be sharper around the true class on the simplex, and the KL-divergence term can be viewed as a regularizer to prevent overconfident prediction, and $\lambda$ is the hyper-parameter to balance the trade-off.

\paragraph{Validation for Uncertainty Learning} \label{sec:validation}

Validation with early stopping is a commonly used technique in supervised learning to train a model with desired generalization performance, i.e., stop training when the error evaluated on the validation set starts increasing. However, we observe that the standard validation method does not work well for uncertainty learning. One possible explanation is that model achieves the highest accuracy when validation loss is small, but may not achieve the best UQ performance, i.e., the model can be overconfident. To this end, we propose a simple and effective validation approach specifically for uncertainty learning. Instead of monitoring the validation cross-entropy loss, we evaluate a specific uncertainty quantification performance metric. For example, we create another noisy validation set for the OOD task by adding noise to the original validation samples and treating such noisy validation samples as OOD samples (more details are provided in the Appendix~\ref{ood-dataset-appendix}). We evaluate the uncertainty score $\vu\left(\tilde{\vy}\right)$ on both the validation set and noisy validation set and stop the meta-model training when the OOD detection performance achieves its maximum based on some predefined metrics, e.g., AUROC score. Unlike most existing literature using additional training data to help achieve desired performance~\citep{hendrycks2018deep, kristiadi2021learnable, malinin2018predictive}, we nonetheless do not require any additional data for training the meta-model.

\subsection{Uncertainty Metrics} \label{sec:metric}
In this section, we show that our meta-model has the desired behavior to quantify different uncertainties and how they can be used in various applications.


\textbf{Total Uncertainty.} 
Total uncertainty, also known as predictive uncertainty, is a combination of epistemic uncertainty and aleatoric uncertainty. The total uncertainty is often used for misclassification detection problems, where the misclassified samples are viewed as in-distribution hard samples. There are two standard ways to measure total uncertainty: (1) \textbf{Entropy} (Ent): The Shannon entropy of expected categorical label distribution over the Dirichlet distribution, i.e., $\gH\left(P(y|\Phi(\vx); \vw_g) \right) = \gH\left(\E_{P(\boldsymbol{\pi}|\Phi(\vx); \vw_g)}[P(y|\boldsymbol{\pi})]\right)
$; (2) \textbf{Max Probability} (MaxP): The probability of the predicted class in label distribution, i.e., $\max_c P(y=c|\Phi(\vx); \vw_g)$.

\textbf{Epistemic Uncertainty.}
The epistemic uncertainty quantifies the uncertainty when the model has insufficient knowledge of a prediction, e.g., 
the case of an unseen data sample. The epistemic uncertainty is especially useful in OOD detection problems. When the meta-model encounters an unseen sample during testing, it will output a high epistemic uncertainty score due to a lack of knowledge. We define three metrics to measure the epistemic uncertainties. (1) \textbf{Differential Entropy} (Dent): Differential entropy measures the entropy of the Dirichlet distribution, a large differential entropy corresponds a more spread Dirichlet distribution, i.e., $\gH\left(P(\boldsymbol{\pi}| \Phi(\vx_i); \vw_g) \right) = -\int P(\boldsymbol{\pi}| \Phi(\vx_i); \vw_g)\cdot \log{P(\boldsymbol{\pi}| \Phi(\vx_i); \vw_g)} d\boldsymbol{\pi}$. (2) \textbf{Mutual Information} (MI): Mutual Information is the difference between the Entropy (measures total uncertainty) and the expected entropy of the categorical distribution sampled from the Dirichlet distribution (approximates aleatoric uncertainty), i.e., $\gI\left(y,\boldsymbol{\pi}|\Phi(\vx_i) \right) = \gH\left(\E_{P(\boldsymbol{\pi}|\Phi(\vx); \vw_g)} [P(y|\boldsymbol{\pi})]\right)- \E_{P(\boldsymbol{\pi}|\Phi(\vx); \vw_g)}[\gH\left(P(y|\boldsymbol{\pi})\right)]$.  (3) \textbf{Precision}(Prec): The precision is the summation of the Dirichlet distribution concentration parameter $\boldsymbol{\alpha}$, larger value corresponds to sharper distribution and higher confidence, i.e., $\alpha_0 = \sum_{c=1}^k \alpha_c$.

\section{Experiment Results} \label{sec:exp}
In this section, we will demonstrate the strong empirical performance of our proposed meta-model-based uncertainty learning method: first, We introduce the UQ applications; then we describe the experiment settings; next, we present the main results of the three aforementioned uncertainty quantification applications; and finally, we discuss our takeaways. More experiment results and implementation details are given in the Appendix~\ref{setup-appendix} and Appendix~\ref{results-appendix}.

\subsection{Uncertainty Quantification Applications} \label{sec:app}
We primarily focus on three applications that can be tackled using our proposed meta-model approach: (1) \textbf{Out of domain data detection.}
Given a base-model $\vh$ trained using data sampled from the distribution $P^B_{Z}$. We use the same base-model training set to train the meta-model, i.e., $\mathcal{D}_{B} = \mathcal{D}_{M}$. During testing, there exists some unobserved out-of-domain data from another distribution $P^{ood}_{Z}$. The meta-model is expected to identify the out-of-distribution input sample based on epistemic uncertainties.  (2) \textbf{misclassification Detection.}
Instead of detecting whether a testing sample is out of domain, the goal here is to identify the failure or success of the meta-model prediction at test time using total uncertainties. (3) \textbf{Trustworthy Transfer Learning.}
In transfer learning, there exists a pretrained model trained using source task data $\mathcal{D}_{s}$ sampled from source distribution $P^s_{Z}$, and the goal is to adapt the source model to a target task using target data $\mathcal{D}_{t}$ sampled from target distribution $P^t_{Z}$. Most existing transfer learning approaches only focus on improving the prediction performance of the transferred model, but ignore its UQ performance on the target task. Our meta-model method can be utilized to address this problem, i.e, given pretrained source model $\vh^{s} \circ \Phi^{s}$, the meta-model can be efficiently trained using target domain data by $\vg^t = \argmin_{\vg} \gL_E(\vg \circ \Phi^s,\gD_t)$.

\begin{table*}[t]
  \begin{center}
  \scriptsize
  \captionsetup{font=small}
  \caption{\textbf{OOD Detection AUROC score.} MI, Dent, and Prec stand for different epistemic uncertainty metrics, i.e., Mutual Information, Differential Entropy, and precision. Settings stand for post-hoc or traditional, i.e., training the entire model from scratch. Additional Data stands for if using additional training data or not.}
  \vspace{-2em}
  \begin{tabular}{lllllllll}
    \\
    \toprule
    \textbf{ID Data}\ \&\ \textbf{Model} & \textbf{Methods} & \textbf{Settings} & \textbf{Additional Data} & \textbf{Omniglot}& \textbf{FMNIST} & \textbf{KMNIST}& \textbf{CIFAR10} &\textbf{Corrupted} \\
    \hline
    MNIST & Base Model(Ent) & Traditional & No & 98.9$\pm$0.5 & 97.8$\pm$0.8 & 95.8$\pm$0.8 & 99.4$\pm$0.2 & 99.5$\pm$0.3  \\
    LeNet & Base Model(MaxP) & Traditional & No & 98.7$\pm$0.6 & 97.6$\pm$0.8 & 95.6$\pm$0.8 & 99.3$\pm$0.2 & 99.4$\pm$0.4  \\
              & Whitebox & Post-hoc & Yes & 98.5$\pm$0.3 & 97.7$\pm$0.6 & 96.0$\pm$0.2 & 99.5$\pm$0.1 & 99.5$\pm$0.1  \\
              & LULA & Post-hoc & Yes & 99.8$\pm$0.0 & 99.4$\pm$0.0 & \textcolor{violet}{\textbf{99.3}$\pm$0.1} & 99.9$\pm$0.0 & 99.6$\pm$0.1  \\
    \hline
              & \textbf{Ours(Ent)} & Post-hoc & No & 99.7$\pm$0.1 & 99.5$\pm$0.2 & 98.2$\pm$0.3 &  \textcolor{violet}{\textbf{100.0}$\pm$0.0} & \textcolor{violet}{\textbf{100.0}$\pm$0.0} \\
              & \textbf{Ours(MaxP)} & Post-hoc & No & 99.3$\pm$0.2 & 99.3$\pm$0.2 & 98.0$\pm$0.2 &  \textcolor{violet}{\textbf{100.0}$\pm$0.0} & \textcolor{violet}{\textbf{100.0}$\pm$0.0} \\
              & \textbf{Ours(MI)} & Post-hoc & No & \textcolor{violet}{\textbf{99.9}$\pm$0.0} & \textcolor{violet}{\textbf{99.6}$\pm$0.2} &97.7$\pm$0.4 &  \textcolor{violet}{\textbf{100.0}$\pm$0.0} & \textcolor{violet}{\textbf{100.0}$\pm$0.0} \\
              & \textbf{Ours(Dent)} & Post-hoc & No & 99.8$\pm$0.0 & 99.5$\pm$0.2 & 97.6$\pm$0.4 & \textcolor{violet}{\textbf{100.0}$\pm$0.0} & \textcolor{violet}{\textbf{100.0}$\pm$0.0} \\
              & \textbf{Ours(Prec)} & Post-hoc & No & \textcolor{violet}{\textbf{99.9}$\pm$0.0} & \textcolor{violet}{\textbf{99.5}$\pm$0.2} &97.7$\pm$0.5 &  \textcolor{violet}{\textbf{100.0}$\pm$0.0} & \textcolor{violet}{\textbf{100.0}$\pm$0.0} \\
    \toprule
    \textbf{ID Data}\ \&\ \textbf{Model} & \textbf{Methods} & \textbf{Settings} & \textbf{Additional Data} & \textbf{SVHN}& \textbf{FMNIST} & \textbf{LSUN}& \textbf{TinyImageNet} &\textbf{Corrupted} \\
    \hline
    CIFAR10 & Base Model(Ent) & Traditional & No & 86.4$\pm$4.6 & 90.8$\pm$1.3 & 89.0$\pm$0.5 & 87.5$\pm$1.1 & 85.9$\pm$8.2  \\
    VGG16   & Base Model(MaxP) & Traditional & No & 86.3$\pm$4.4 & 90.4$\pm$1.2 & 88.7$\pm$0.5 & 87.3$\pm$1.1 & 85.7$\pm$8.1   \\
              & Whitebox & Post-hoc & Yes & 96.9$\pm$0.9 & 95.2$\pm$1.2 & 89.3$\pm$2.2 & 88.9$\pm$2.5 & 96.4$\pm$1.0   \\
              & LULA & Post-hoc & Yes & 97.1$\pm$1.7 & 94.3$\pm$0.0 & 92.8$\pm$0.1 & 90.0$\pm$0.0 & 97.7$\pm$2.0  \\
    \hline
              & \textbf{Ours(Ent)} & Post-hoc & No & 96.3$\pm$3.0 &89.0$\pm$5.2 & 89.6$\pm$3.4 &  89.4$\pm$3.5& 95.9$\pm$4.3 \\
              & \textbf{Ours(MaxP)} & Post-hoc & No & 95.6$\pm$3.6 & 87.8$\pm$4.4 &89.1$\pm$2.4 &  88.2$\pm$2.6 & 94.0$\pm$7.3 \\
              & \textbf{Ours(MI)} & Post-hoc & No & \textcolor{violet}{\textbf{100.0$\pm$0.0}} & \textcolor{violet}{\textbf{98.8$\pm$0.5}} & 95.2$\pm$0.9 &  \textcolor{violet}{\textbf{98.1$\pm$0.3}} & \textcolor{violet}{\textbf{100.0$\pm$0.0}} \\
              & \textbf{Ours(Dent)} & Post-hoc & No & \textcolor{violet}{\textbf{100.0$\pm$0.0}} & 98.4$\pm$0.9 & \textcolor{violet}{\textbf{95.7}$\pm$0.8} &  97.7$\pm$0.5 & \textcolor{violet}{\textbf{100.0$\pm$0.0}} \\
               & \textbf{Ours(Prec)} & Post-hoc & No & \textcolor{violet}{\textbf{100.0$\pm$0.0}} & \textcolor{violet}{\textbf{98.8$\pm$0.5}} & 95.1$\pm$0.5 &  \textcolor{violet}{\textbf{98.1$\pm$0.3}} & \textcolor{violet}{\textbf{100.0$\pm$0.0}} \\
    \toprule
    CIFAR100 & Base Model(Ent) & Traditional & No & 76.2$\pm$5.2 & 77.8$\pm$2.4 & 80.1$\pm$0.5 & 79.7$\pm$0.3 & 65.8$\pm$11.4  \\
    WideResNet   & Base Model(MaxP) & Traditional & No & 73.9$\pm$4.3 & 76.4$\pm$2.3 & 78.7$\pm$0.5 & 78.0$\pm$0.2 & 63.8$\pm$10.4   \\
              & Whitebox & Post-hoc & Yes & 89.0$\pm$0.7 & 82.4$\pm$1.1 & 80.5$\pm$0.7 & 79.0$\pm$1.1 & 83.1$\pm$1.6   \\
              & LULA & Post-hoc & Yes & 84.2$\pm$1.0 & 83.2$\pm$0.1 & 79.6$\pm$0.3 & 78.5$\pm$0.1 & 80.6$\pm$1.0  \\
    \hline
              & \textbf{Ours(Ent)} & Post-hoc & No & 92.6$\pm$2.0 & 80.8$\pm$3.0 & 81.1$\pm$1.2 &  84.9$\pm$1.2& 85.6$\pm$3.9 \\
              & \textbf{Ours(MaxP)} & Post-hoc & No & 88.6$\pm$3.2 & 78.4$\pm$3.0 & 79.7$\pm$0.6 &  82.4$\pm$0.6& 79.3$\pm$4.5 \\
              & \textbf{Ours(MI)} & Post-hoc & No & 94.3$\pm$1.0 & \textcolor{violet}{\textbf{84.4}$\pm$1.8} & 81.9$\pm$4.7 &  85.5$\pm$3.6 & \textcolor{violet}{\textbf{90.8}$\pm$3.3} \\
              & \textbf{Ours(Dent)} & Post-hoc & No & 93.3$\pm$1.4 & 84.0$\pm$2.3 & 79.5$\pm$3.0 &  84.6$\pm$2.8& 89.8$\pm$2.6 \\
              & \textbf{Ours(Prec)} & Post-hoc & No & \textcolor{violet}{\textbf{94.4}$\pm$1.0} & \textcolor{violet}{\textbf{84.4}$\pm$1.7} & \textcolor{violet}{\textbf{82.1}$\pm$4.9} &  \textcolor{violet}{\textbf{85.6}$\pm$3.6}&
              \textcolor{violet}{\textbf{90.8}$\pm$3.4} \\
    \toprule
  \end{tabular}
  \label{table:OOD}
  \end{center}
  \vspace{-2em}
\end{table*}

\subsection{Settings}\label{sec:setting}
\paragraph{Benchmark.}
For both OOD detection and misclassification detection tasks, we employ three standard datasets to train the base model and the meta-model: MNIST, CIFAR10, and CIFAR100. For each dataset, we use different base-model structures, i.e., LeNet for MNIST, VGG-16~\citep{simonyan2014very} for CIFAR10, and WideResNet-16~\citep{zagoruyko2016wide} for CIFAR100. For LeNet and VGG-16, the meta-model uses extracted feature after each pooling layer, and for WideResNet-16, the meta-model uses extracted feature after each residual black. In general, the total number of intermediate features is less than 5 to ensure computational efficiency. For the OOD task, we consider five different OOD datasets for evaluating the OOD detection performance: Omiglot, FashionMNIST, KMNIST, CIFAR10, and corrupted MNIST as outliers for the MNIST dataset; SVHN, FashionMNIST, LSUN, TinyImageNet, and corrupted CIFAR10 (CIFAR100) as outliers for CIFAR10 (CIFAR100) dataset. For the trustworthy transfer learning task, we use the ResNet-50 pretrained on ImageNet as our pretrained source domain model and adapt the source model to the two target tasks, STL10 and CIFAR10, by training the meta-model.

\paragraph{Baselines.}
For OOD and misclassification tasks, except the naive base-model trained with cross-entropy loss, we mainly compare with the existing post-hoc UQ methods as baselines: (1) The meta-model based method (Whitebox)~\citep{chen2019confidence}; (2) The post-hoc uncertainty quantification using Laplace Approximation (LULA)~\citep{kristiadi2021learnable}. In order to further validates our strong empirical performance, we also compare with other SOTA intrinsic UQ methods in the Appendix~\ref{results-appendix}: (1) The standard Bayesian method Monte-Carlo Dropout~\citep{gal2016dropout}; (2) The Dirichlet network with variational inference (Be-Bayesian)~\citep{joo2020being}; (3) The posterior network with density estimation~\citep{charpentier2020posterior}; (4) The robust OOD detection method ALOE~\citep{chen2020robust}.

For the trustworthy transfer learning task, since there is no existing work designed for this problem, we compare our method with two simple baselines: (1) Fine-tune the last layer of the source model. (2) Train our proposed meta-model on top of the source model using standard cross-entropy loss.

\paragraph{Performance Metrics.}
We evaluate the UQ performance by measuring the area under the ROC curve (AUROC) and the area under the Precision-Recall curve (AUPR). The results are averaged over five random trails for each experiment. For the OOD task, we consider the in-distribution test samples as the negative class and the outlier samples as the positive class. For the misclassification task, we consider correctly classified test samples as the negative class and miss-classified test samples as the positive class.

\subsection{OOD Detection}\label{sec:OOD}
The OOD detection results for the three benchmark datasets, MNIST, CIFAR10, and CIFAR100, are shown in Table~\ref{table:OOD}. Additional baseline comparisons are provided in Table~\ref{table:OOD-appendix} in Appendix. Our proposed Dirichlet meta-model method consistently outperforms all the baseline methods in terms of AUROC score (AUPR results are shown in Appendix), including the recent proposed SOTA post-hoc uncertainty learning method LULA. We also evaluate the performance of all the uncertainty metrics defined in section~\ref{sec:metric}, as it can be observed that compared to total uncertainty (Ent and MaxP), epistemic uncertainties (MI, Dent, Prec) can achieve better UQ performance for the OOD detection task. Moreover, our proposed method does not require additional data to train the meta-model. In contrast, Whitebox requires an additional validation set to train the meta-model, and LULA also needs an additional OOD dataset during training to distinguish the in-distribution samples and outliers, which imposes practical limitations.

\subsection{Misclassification Detection}\label{sec:Miss}
The misclassification detection results for the three benchmark datasets, MNIST, CIFAR10, and CIFAR100, are shown in Table~\ref{table:miss-class}. Additional baseline comparisons are provided in Table~\ref{table:miss-class-appendix}. LULA turns out to be a strong baseline for the misclassification detection task. Although our proposed method performs slightly worse than LULA in terms of the AUROC, it outperforms all the baselines in terms of AUPR.

\begin{table*}[t]
  \begin{center}
  \scriptsize
  \captionsetup{font=small}
  \caption{\textbf{Misclassification Results.} Ent and MaxP stand for Entropy and Max Probability, respectively. Settings stand for post-hoc or traditional, i.e., training the entire model from scratch.}
  \vspace{-1.5em}
  \begin{tabular}{llllllllll}
    \\
    \toprule
    \textbf{Methods} & \textbf{Settings} & \textbf{Metric} & \textbf{MNIST}& \makecell{\textbf{CIFAR 10}} & \makecell{\textbf{CIFAR 100}}& \textbf{Metric} &  \textbf{MNIST}& \makecell{\textbf{CIFAR 10}} & \makecell{\textbf{CIFAR 100}}\\
    \hline
    Base Model(Ent)  & Traditional & AUROC & 96.7$\pm$0.9 & 92.1$\pm$0.2 & 87.2$\pm$0.2 & AUPR & 37.4$\pm$4.0 & 47.5$\pm$1.2 & 67.0$\pm$1.2  \\
    Base Model(MaxP) & Traditional & AUROC & 96.7$\pm$0.9 & 92.1$\pm$0.2 & 86.8$\pm$0.2 & AUPR & 39.4$\pm$3.6 & 46.6$\pm$1.1 & 65.7$\pm$1.0  \\
    Whitebox &Post-hoc & AUROC & 94.9$\pm$0.2 & 90.2$\pm$0.1 & 80.3$\pm$0.1 & AUPR & 30.4$\pm$0.3 & 45.7$\pm$0.2 & 52.5$\pm$0.3  \\
    LULA &Post-hoc & AUROC & \textcolor{violet}{\textbf{98.8}$\pm$0.1} & \textcolor{violet}{\textbf{94.5}$\pm$0.0} & \textcolor{violet}{\textbf{87.5}$\pm$0.1} & AUPR & 40.7$\pm$4.2 & 47.3$\pm$0.7 & 66.0$\pm$0.4  \\
    \hline
    \textbf{Ours(Ent)} &Post-hoc & AUROC  & 96.9$\pm$0.6 & 91.1$\pm$0.2 & 83.4$\pm$0.1 & AUPR & 35.6$\pm$4.5 & 50.0$\pm$3.1 & 66.3$\pm$0.4  \\
    \textbf{Ours(MaxP)} &Post-hoc & AUROC  & 97.4$\pm$0.4 & 92.2$\pm$0.7 & 85.8$\pm$0.2 & AUPR & \textcolor{violet}{\textbf{44.5}$\pm$5.1} & \textcolor{violet}{\textbf{54.2}$\pm$3.2} & \textcolor{violet}{\textbf{68.2}$\pm$0.5}  \\
    \toprule
  \end{tabular}
  \label{table:miss-class}
  \end{center}
  \vspace{-1.2em}
\end{table*}

\begin{table*}[t]
  \begin{center}
  \scriptsize
  \captionsetup{font=small}
  \caption{\textbf{Trustworthy Transfer Learning Results.} Ent, MaxP, MI, and Dent stand for different uncertainty measurements, i.e., Entropy, Max Probability, Mutual Information, and Differential Entropy, respectively. We use ResNet-50 pretrained with ImageNet as the source model and FashionMNIST as OOD samples.}
  \vspace{-1.5em}
  \begin{tabular}{lllllllll}
    \\
    \toprule
    \textbf{Methods} &\textbf{Target} &\textbf{Test Acc} & \textbf{AUROC} & \textbf{AUPR}& \textbf{Target} &\textbf{Test Acc}& \textbf{AUROC} &  \textbf{AUPR} \\
    \hline
    FineTune(Ent)  &\textbf{STL10} &48.1$\pm$0.5  & 89.2$\pm$0.9 & 89.3$\pm$1.2 & \textbf{CIFAR10} &65.0$\pm$0.4 &74.8$\pm$1.6 &71.6$\pm$1.8  \\
    FineTune(MaxP) &               &48.1$\pm$0.5  & 81.8$\pm$1.2 & 83.2$\pm$1.7 &                  &65.0$\pm$0.4 &72.7$\pm$1.4&69.4$\pm$1.5  \\
    \hline
    CrossEnt Loss(Ent)  &    &48.0$\pm$0.2  & 88.9$\pm$1.0 & 87.4$\pm$1.5 &  &86.3$\pm$0.1 &85.0$\pm$ 1.0& 82.2$\pm$1.9  \\
    CrossEnt Loss(MaxP) &               &48.0$\pm$0.2  & 84.9$\pm$0.7 & 84.7$\pm$0.8 &                  &86.3$\pm$0.1 &83.1$\pm$0.9& 79.0$\pm$1.6  \\
    \hline
    \textbf{Ours(Ent)} &          &47.2$\pm$0.3 & \textcolor{violet}{\textbf{91.8$\pm$0.8}} & \textcolor{violet}{\textbf{91.3$\pm$0.7}} &  &86.6$\pm$0.3 &89.9$\pm$1.3&88.8$\pm$1.5\\
    \textbf{Ours(MaxP)}&          &47.2$\pm$0.3 & 87.3$\pm$1.7 & 87.9$\pm$1.5 &  &86.6$\pm$0.3 &87.6$\pm$1.6&85.5$\pm$2.0  \\
     \textbf{Ours(MI)} &        &47.2$\pm$0.3 & 90.2$\pm$2.0 & 88.7$\pm$2.8 &  &86.6$\pm$0.3 &90.8$\pm$0.9&89.9$\pm$1.2\\
    \textbf{Ours(Dent)}&        &47.2$\pm$0.3 & 91.7$\pm$1.0 & 90.3$\pm$1.7 &  &86.6$\pm$0.3 &\textcolor{violet}{\textbf{92.0$\pm$0.8}}& \textcolor{violet}{\textbf{91.1$\pm$0.8}}\\
    \toprule
  \end{tabular}
  \label{table:transfer}
  \end{center}
  \vspace{-1.5em}
\end{table*}

\subsection{Trustworthy Transfer Learning}\label{sec:transfer}
We use ImageNet pretrained ResNet-50 as our source domain base model and adapt the pretrained model to the target task by training the meta-model using the target domain training data. Unlike traditional transfer learning, which only focuses on testing prediction accuracy on the target task, we also evaluate the UQ ability of the meta-model in terms of OOD detection performance. We use FashionMNIST as OOD samples for both target tasks STL10 and CIFAR10 and evaluate the AUROC score. The results are shown in Table~\ref{table:transfer}. Our proposed meta-model method can achieve comparable prediction performance to the baseline methods and significantly improve the OOD detection performance, which is crucial in trustworthy transfer learning.

\subsection{Discussion}\label{sec:ablation}

In this section, we further investigate our proposed method through an ablation study using the CIFAR10 OOD task. Based on our insights and the empirical results, we conclude the following four key factors in the success of our meta-model based method:

\textbf{Feature Diversity.}
We replace our proposed meta-model structure with a simple linear classifier attached to only the final layer. The ablation results are shown in Table~\ref{table:ablation-structure} as ``\textbf{Linear-Meta}''. It can be observed here and in Figure~\ref{fig: toy_problem} that the performance degrades without using features from all intermediate layers, which further justifies the importance of feature diversity and the effectiveness of our meta-model structure. 


\textbf{Dirichlet Technique.}
Instead of using a meta-model to parameterize a Dirichlet distribution, we train the meta-model using the standard cross-entropy loss, which simply outputs a categorical label distribution. The ablation results are shown in Table~\ref{table:ablation-structure} as ``\textbf{Cross-Ent}''. It can be shown that performance degrades again because it cannot quantify epistemic uncertainty, which justifies the effectiveness of using Bayesian techniques.

\textbf{Validation for Uncertainty Learning.}
We retrain the last layer of the base model using the cross-entropy loss with the proposed validation trick. The results are shown in Table~\ref{table:ablation-structure} as ``\textbf{LastLayer}''. It turns out even such a naive method can achieve improved performance compared to the base model, which further justifies the effectiveness of the post-hoc uncertainty learning setting, i.e., the benefit of solely focusing on UQ performance at the second stage. This interesting observation inspires us to conjecture that efficiently retraining the classifier of the base model at the second stage will lead to better UQ performance. A theoretical investigation of this observation can be interesting for future work.

\textbf{Data Efficiency.}
Instead of using all the training samples, we randomly choose only $10\%$ samples to train the meta-model. The results are shown in Table~\ref{table:ablation-structure} as ``$10\%$\textbf{data}''. It can be observed that our meta-model requires a small amount of data to achieve comparable performance due to the smaller model complexity. Therefore, our proposed method is also more computationally efficient than the approaches that retrain the whole model from scratch.

\begin{table}[h]
  \begin{center}
  \scriptsize
  \captionsetup{font=small}
  \caption{\textbf{Ablation Study of Meta-model (CIFAR10 AUROC score).} The results are reported as mean over five experiment trails. Error bars are provided in Table~\ref{table:ablation-structure-appendix} in the Appendix.}
  \vspace{-1.5em}
  \begin{tabular}{lllllll}
    \\
    \toprule
    \textbf{Methods} & \textbf{SVHN}& \textbf{FMNIST} & \textbf{LSUN}& \textbf{TIM} &\textbf{Corrupted} \\
    \hline
     Base Model(Ent)  & 86.4 & 90.8 & 89.0 & 87.5 & 85.9  \\
    Base Model(MaxP)  & 86.3 & 90.4 & 88.7 & 87.3 & 85.7   \\
    \hline
            \textbf{Linear-Meta(Ent)}  & 90.4 & 91.3 & 91.5 & 89.5 & 90.4 \\
            \textbf{Linear-Meta(MaxP)}  & 90.1 & 91.5 & 91.4 & 89.6 & 90.1 \\
            \textbf{Linear-Meta(MI)}  & 90.6 & 90.1 & 91.2 & 88.8 & 90.5 \\
            \textbf{Linear-Meta(Dent)}  & 90.4 & 90.7 & 91.4 & 89.2 & 90.3 \\
            \textbf{Linear-Meta(Prec)} & 90.6 & 90.0 & 91.2 & 88.8 & 90.6\\
    \hline
      \textbf{Cross-Ent(Ent)}  & 94.2 & 91.2 & 91.2 &  90.3& 94.7 \\
        \textbf{Cross-Ent(MaxP)}  & 93.3& 91.1 &90.9 &  90.0 & 94.0 \\
    \hline
       \textbf{LastLayer(Ent)}& 93.0& 90.2 & 91.9 &  89.9& 93.1 \\
        \textbf{LastLayer(MaxP)} & 92.9 & 90.5 &91.9&  90.1 & 93.1\\
    \hline        
    \textbf{$10\%$data(Ent)}  & 90.0 &89.1& 88.7 &  88.2& 90.2 \\
              \textbf{$10\%$data(MaxP)} & 90.9 & 88.1&86.9&  86.5& 91.7 \\
              \textbf{$10\%$data(MI)}& 99.9 & 98.1 & 95.4&  97.2 & 99.9 \\
              \textbf{$10\%$data(Dent)}  & 96.7 & 97.4 & 94.3&  95.7 & 96.7\\
               \textbf{$10\%$data(Prec)}  & 99.9& 98.0& 95.4 &  97.3& 99.9 \\
    \hline        
    \textbf{Ours(Ent)}  & 96.3 &89.0& 89.6 &  89.4& 95.9 \\
              \textbf{Ours(MaxP)} & 95.6 & 87.8 &89.1 &  88.2 & 94.0 \\
              \textbf{Ours(MI)}& \textcolor{violet}{\textbf{100.0}} & \textcolor{violet}{\textbf{98.8}} & 95.2 &  \textcolor{violet}{\textbf{98.1}} & \textcolor{violet}{\textbf{100.0}} \\
              \textbf{Ours(Dent)}  & \textcolor{violet}{\textbf{100.0}} & 98.4 & \textcolor{violet}{\textbf{95.7}} &  97.7 & \textcolor{violet}{\textbf{100.0}} \\
            \textbf{Ours(Prec)}  & \textcolor{violet}{\textbf{100.0}} & \textcolor{violet}{\textbf{98.8}} & 95.1 &  \textcolor{violet}{\textbf{98.1}} & \textcolor{violet}{\textbf{100.0}} \\
    \toprule
  \end{tabular}
  \label{table:ablation-structure}
  \end{center}
  \vspace{-1.5em}
\end{table}

\section{Concluding Remarks} \label{sec:conclusion}

We provide a novel solution for the uncertainty quantification problem via our proposed post-hoc uncertainty learning framework and the Dirichlet meta-model approach.
Our method turns out to be both effective and computationally efficient for various UQ applications. We believe our meta-model approach not only has the flexibility to tackle other applications relevant to uncertainty quantification, such as quantifying transfer-ability in transfer learning and domain adaptation, but also can adapt to other model architecture such as transformer and language model. Exploring these potential applications and offering a theoretical interpretation of the meta-model can be interesting future work.

\section*{Acknowledgement}

This work was supported, in part, by the MIT-IBM Watson AI Lab under Agreement No.~W1771646, and NSF under Grant No.~CCF-1816209.

\newpage

\bibliography{aaai23}

\clearpage
\appendix
\section{Derivation of ELBO loss} \label{sec:ELBO-appendix}
The Dirichlet ELBO loss can be fomulated as follows:
\begin{align}
\gL_{\mathrm{VI}}(\vw_g) &=\frac{1}{N_M}\sum_{i=1}^{N_M} \mathbb{E}_{q(\boldsymbol{\pi}| \Phi(\vx_i); \vw_g)}[\log P(y_i \mid \boldsymbol{\pi}, \boldsymbol{x}_i)] \nonumber \\
&- \lambda\cdot\mathrm{KL}\left(q(\boldsymbol{\pi}| \Phi(\vx_i); \vw_g) \| P(\boldsymbol{\pi} | \Phi(\vx_i))\right)
\end{align}
\normalsize
The second term in the summation is simply the KL-divergence between variational distribution parameterized by the meta model and the predefined prior distribution, i.e., $\mathrm{KL}\left( \operatorname{Dir}(\boldsymbol{\pi}| \boldsymbol{\alpha}^{(i)}) \| \operatorname{Dir}(\boldsymbol{\pi} | \boldsymbol{\beta})\right)$. The first term in the summation is the expectation of cross-entropy loss w.r.t the variational Dirichlet distribution, whose closed form can be obtained as follows,
\begin{align}
&\mathbb{E}_{q(\boldsymbol{\pi}| \Phi(\vx_i); \vw_g)}[\log P(y_i \mid \boldsymbol{\pi}, \boldsymbol{x}_i)] \nonumber \\
=& \mathbb{E}_{ \operatorname{Dir}(\boldsymbol{\pi}| \boldsymbol{\alpha}^{(i)})}[\log \boldsymbol{\pi}_{y_i}]\\
=& \int \log \boldsymbol{\pi}_{y_i} \operatorname{Dir}(\boldsymbol{\pi}| \boldsymbol{\alpha}^{(i)}) d\boldsymbol{\pi}\\
=&\int_{0}^{1} \log \boldsymbol{\pi}_{y_i} \operatorname{Beta}(\alpha^{(i)}_{y_i}, \alpha^{(i)}_{0}-\alpha^{(i)}_{y_i}) d\boldsymbol{\pi}_{y_i}\\
=& \int_{0}^{1} \log \boldsymbol{\pi}_{y_i} \frac {\boldsymbol{\pi}_{y_i}^{\alpha^{(i)}_{y_i}-1}(1-\boldsymbol{\pi}_{y_i})^{\alpha^{(i)}_{0}-\alpha^{(i)}_{y_i}-1}} {\operatorname{B}(\alpha^{(i)}_{y_i}, \alpha^{(i)}_{0}-\alpha^{(i)}_{y_i})} d\boldsymbol{\pi}_{y_i}\\
=& \frac {\int_{0}^{1} \frac {d \boldsymbol{\pi}_{y_i}^{\alpha^{(i)}_{y_i}-1}}{d\alpha^{(i)}_{y_i}}(1-\boldsymbol{\pi}_{y_i})^{\alpha^{(i)}_{0}-\alpha^{(i)}_{y_i}-1} d\boldsymbol{\pi}_{y_i}}
{\operatorname{B}(\alpha^{(i)}_{y_i}, \alpha^{(i)}_{0}-\alpha^{(i)}_{y_i})} \\
=& \frac { \frac{d}{d\alpha^{(i)}_{y_i}}  \int_{0}^{1}  \boldsymbol{\pi}_{y_i}^{\alpha^{(i)}_{y_i}-1} (1-\boldsymbol{\pi}_{y_i})^{\alpha^{(i)}_{0}-\alpha^{(i)}_{y_i}-1} d\boldsymbol{\pi}_{y_i}}
{\operatorname{B}(\alpha^{(i)}_{y_i}, \alpha^{(i)}_{0}-\alpha^{(i)}_{y_i})} \\
=& \frac {1} {\operatorname{B}(\alpha^{(i)}_{y_i}, \alpha^{(i)}_{0}-\alpha^{(i)}_{y_i})} \frac {d\operatorname{B}(\alpha^{(i)}_{y_i}, \alpha^{(i)}_{0}-\alpha^{(i)}_{y_i})} {d\alpha^{(i)}_{y_i}}\\
=& \frac {d \log \operatorname{B}(\alpha^{(i)}_{y_i}, \alpha^{(i)}_{0}-\alpha^{(i)}_{y_i})} {d\alpha^{(i)}_{y_i}}\\
=& \frac {d \left(\log \Gamma(\alpha^{(i)}_{y_i})+ \log \Gamma(\alpha^{(i)}_{0}-\alpha^{(i)}_{y_i}) - \log \Gamma(\alpha^{(i)}_{0})\right)} {d\alpha^{(i)}_{y_i}}\\
=& \frac {d \log \Gamma(\alpha^{(i)}_{y_i})}{d\alpha^{(i)}_{y_i}} - \frac {d \log \Gamma(\alpha^{(i)}_{0})}{d\alpha^{(i)}_{0}}\\
=& \psi\left(\alpha^{(i)}_{y_i}\right)-\psi\left(\alpha^{(i)}_{0}\right)
\end{align}  
Therefore, the ELBO loss can be formulated as the closed form as follows,
\begin{align}
\gL_{\mathrm{VI}}(\vw_g) 
&= \frac{1}{N_M}\sum_{i=1}^{N_M} \psi\left(\alpha^{(i)}_{y_i}\right)-\psi\left(\alpha^{(i)}_{0}\right)  \nonumber\\
&- \lambda\cdot \mathrm{KL}\left( \operatorname{Dir}(\boldsymbol{\pi}| \boldsymbol{\alpha}^{(i)}) \| \operatorname{Dir}(\boldsymbol{\pi} | \boldsymbol{\beta})\right).
\end{align}

\newpage
\section{Experiment Setup} \label{setup-appendix}
\subsection{Implementation Details for OOD and Misclassification}
The post-hoc uncertainty learning problem aims to improve the UQ performance of a pretrained base model. First, we generate the pretrained model by training the base model using cross-entropy loss to achieve optimal testing accuracy. The maximum epochs for training LeNet, VGG-16, and WideResNet-16 are set to be 20, 200, and 200, respectively. Then, in the second stage, we freeze the parameter of the pretrained base model and train the meta-model on top of it using the Dirichlet variational loss. The meta-model uses the same training data as the base model, and the maximum epochs for training the meta-model is set to be 50. All the models are optimized using an SGD optimizer. The hyper-parameters for training the base model and meta-model are summarized in Table~\ref{table: hyper}, where $bs$ denotes the batch size, $lr$ denotes the learning rate, $m$ denotes the momentum, $wd$ denotes the weight decay, $\lambda$ is the hyper-parameter to balance the two terms in variational loss, and $\beta$ is the concentration parameter of prior Dirichlet distribution. All experiments are implemented in PyTorch using Titan RTX GPUs with 24 GB memory.
\begin{table*}[t]
  \caption{Hyper-parameter for training the base model and meta-model}
  \centering
  \begin{tabular}{llllllllll}
    \toprule
    Dataset & Model & Epoch & $bs$ & $\lambda$ & $\beta$ & $lr$ & $m$ & $wd$\\
    \midrule
    MNIST & LeNet-Base & 20 & 128 & / & / & 0.01 & 0.9 & $5\times10^{-4}$ \\
    \midrule
    MNIST & LeNet-Meta & 50 & 128 & $10^{-1}$ & 1 & 0.1 & 0.9 & $5\times10^{-4}$ \\
    \midrule
    CIFAR10 & VGG16-Base & 200 & 128 & / & / & 0.1 & 0.9 & $1\times10^{-4}$ \\
    \midrule
    CIFAR10 & VGG16-Meta & 50 & 128 & $10^{-3}$ & 1 & 0.001 & 0.9 & $1\times10^{-4}$ \\
    \midrule
    CIFAR100 & WideResNet-Base & 200 & 128 & / & / & 0.1 & 0.9 & $1\times10^{-4}$ \\
    \midrule
    CIFAR100 & WideResNet-Meta & 50 & 128 & $10^{-3}$ & 1 & 0.1 & 0.9 & $1\times10^{-4}$ \\
    \bottomrule
  \end{tabular}
  \label{table: hyper}
\end{table*}
\subsection{Implementation Details of Trustworthy Trnasfer Learning}
We download the pretrained ResNet-50 trained on ImageNet as the base-model. Similarly, we freeze the parameter of the pretrained model and train the meta-model on top of it using the training data of the target task. All the models are optimized using an SGD optimizer. The hyper-parameters for training the meta-model are summarized in Table~\ref{table: hyper-transfer}, where $bs$ denotes the batch size, $lr$ denotes the learning rate, $m$ denotes the momentum, $wd$ denotes the weight decay, $\lambda$ is the hyper-parameter to balance the two terms in variational loss, and $\beta$ is the concentration parameter of prior Dirichlet distribution. 
\begin{table*}[t]
  \caption{Hyper-parameter}
  \centering
  \begin{tabular}{llllllllll}
    \toprule
    Dataset & Model & Epoch & $bs$ & $\lambda$ & $\beta$ & $lr$ & $m$ & $wd$\\
    \midrule
    STL10 & ResNet50-Meta & 50 & 128 & $10^{-3}$ & 1 & 0.01 & 0.9 & $1\times10^{-4}$ \\
    \midrule
    CIFAR10 & ResNet50-Meta & 50 & 128 & $10^{-3}$ & 1 & 0.01 & 0.9 & $1\times10^{-4}$ \\
    \bottomrule
  \end{tabular}
  \label{table: hyper-transfer}
\end{table*}

\subsection{Meta-Model Structure}
The high-level description of the meta-model structure is provided in~\ref{subsec: model}. More specifically, all the linear layers $\vg_{i}$ and $\vg_{c}$ consist of three elements: a fully-connected layer, ReLU activation function, and Max-Pooling. Each $\vg_{i}$ has multiple fully-connected layers, each is followed by a ReLU and a Max-Pooling, each fully-connected layer reduces the input feature dimension to half size, and the output meta-feature of $\vg_{i}$ has the dimension as the class number, e.g., 10 for CIFAR10. The final linear layer $\vg_{c}$ is a single fully-connected layer that takes the concatenation of all the meta-feature and outputs the concentration parameter $\alpha$.

\subsection{Training Time Complexity}
Our meta-model-based approach is much more efficient than traditional uncertainty quantification approaches due to the simpler structure and faster convergence speed. To quantitatively demonstrate such efficiency, we measure the wall-clock time of training the meta-model in seconds (on a single Titan RTX GPU) as follows. The training time of training VGG16 model on CIFAR10 dataset is 66.5s for five epochs; The training time of training WideResNet model on CIFAR100 dataset is 241.9s for ten epochs. The training time of the meta-model can be negligible compared to those approaches training the entire base model from scratch (usually taking several hours).

\subsection{Validation for Uncertainty Learning}
We use the proposed validation trick discussed in~\ref{sec:validation} to perform early stopping in the training of the meta-model. We randomly pick $\%20$ training data as our validation set. For the OOD detection task, we create the noisy validation set by by applying various kinds of noise and perturbation to the original images, including permuting the pixels, applying Gaussian blurring, and contrast re-scaling. For the misclassification task, we directly use the validation set to evaluate the misclassification detection performance with the correctly classified and miss-classified validation samples.

\subsection{Description of OOD datasets} \label{ood-dataset-appendix}
For OOD detection task, we use the testing set as in-domain dataset, and make sure the out of domain dataset has the same number of samples (10000 samples) as in-domain dataset. Different dataset input images are resized to 32x32 to ensure they have the same size, and all gray-scale images are converted into three-channel images. We use the following datasets as OOD samples for the OOD detection task:
\begin{itemize}
    \item \textbf{Omniglot} contains 1632 hand-written characters taken from 50 different alphabets. We randomly pick 10000 images from its testing set as OOD samples for MNIST.
    \item \textbf{Fashion-MNIST} is a dataset of Zalando's article images. We use the testing set with 10000 images as OOD samples for both MNIST and CIFAR.
    \item \textbf{KMNIST} contains handwritten characters from the Japanese Kuzushiji texts. We use the testing set with 10000 images as OOD samples for MNIST.
    \item \textbf{SVHN} contains images of house numbers taken from Google Street View. We use the testing set with 10000 images as OOD samples for CIFAR.
    \item \textbf{LSUN} The Large-scale Scene UNderstanding dataset(LSUN) is a dataset of different objects taken from 10 different scene categories. We use the images from the classroom categories and randomly sample 10000 training images as OOD samples for CIFAR.
    \item \textbf{TIM} TinyImageNet(TIM) is a subset of the ImageNet dataset, and we use the validation set with 10000 images as OOD samples for CIFAR
    \item \textbf{Corrupted} is an artificial dataset generated by perturbing the original testing image using Gaussian blurring, pixel permutation, and contrast re-scaling.
\end{itemize}

\section{Additional Experiment Results} \label{results-appendix}
\subsection{OOD Detection}
In the following, we compare our proposed method with several SOTA uncertainty quantification methods with traditional settings on the OOD detection task. The results are shown in~Table \ref{table:OOD-appendix}.  our proposed method can still outperform these methods.
We provide additional OOD detection results in terms of AUPR score as shown in~Table \ref{table: OOD-aupr}.
\begin{table*}[t]
  \begin{center}
  \scriptsize
  \captionsetup{font=small}
  \caption{\textbf{OOD Detection AUROC score.} MI, Dent, and Prec stand for different epistemic uncertainty metrics, i.e., Mutual Information, Differential Entropy, and precision. Settings stand for post-hoc or traditional, i.e., training the entire model from scratch. Additional Data stands for if using additional training data or not.}
  \vspace{-1.5em}
  \begin{tabular}{lllllllll}
    \\
    \toprule
    \textbf{ID Data}\ \&\ \textbf{Model} & \textbf{Methods} & \textbf{Settings} & \textbf{Additional Data} & \textbf{Omniglot}& \textbf{FMNIST} & \textbf{KMNIST}& \textbf{CIFAR10} &\textbf{Corrupted} \\
    \hline
    MNIST & Base Model(Ent) & Traditional & No & 98.9$\pm$0.5 & 97.8$\pm$0.8 & 95.8$\pm$0.8 & 99.4$\pm$0.2 & 99.5$\pm$0.3  \\
    LeNet & Base Model(MaxP) & Traditional & No & 98.7$\pm$0.6 & 97.6$\pm$0.8 & 95.6$\pm$0.8 & 99.3$\pm$0.2 & 99.4$\pm$0.4  \\
              & MCDropout & Traditional & No & 98.2$\pm$0.1 & 98.1$\pm$0.4 & 92.9$\pm$0.3 & 99.3$\pm$0.4 & 98.7$\pm$0.2  \\
              & BeBayesian & Traditional & No & 99.2$\pm$0.4 & 98.5$\pm$0.5 & 96.1$\pm$0.9 & 99.5$\pm$0.2 & 99.7$\pm$0.2  \\
              & PostNet & Traditional & No & 99.0$\pm$0.5 & 99.1$\pm$0.5 & 99.3$\pm$0.3 & 99.0$\pm$0.2 & 98.9$\pm$0.7  \\
              & ALOE & Traditional & Yes & 100.0$\pm$0.1 & 99.0$\pm$0.3 & 96.7$\pm$0.6 & 99.8$\pm$0.1 & 100.0$\pm$0.0  \\
    \hline
              & \textbf{Ours(Ent)} & Post-hoc & No & 99.7$\pm$0.1 & 99.5$\pm$0.2 & 98.2$\pm$0.3 &  \textcolor{violet}{\textbf{100.0}$\pm$0.0} & \textcolor{violet}{\textbf{100.0}$\pm$0.0} \\
              & \textbf{Ours(MaxP)} & Post-hoc & No & 99.3$\pm$0.2 & 99.3$\pm$0.2 & 98.0$\pm$0.2 &  \textcolor{violet}{\textbf{100.0}$\pm$0.0} & \textcolor{violet}{\textbf{100.0}$\pm$0.0} \\
              & \textbf{Ours(MI)} & Post-hoc & No & \textcolor{violet}{\textbf{99.9}$\pm$0.0} & \textcolor{violet}{\textbf{99.6}$\pm$0.2} &97.7$\pm$0.4 &  \textcolor{violet}{\textbf{100.0}$\pm$0.0} & \textcolor{violet}{\textbf{100.0}$\pm$0.0} \\
              & \textbf{Ours(Dent)} & Post-hoc & No & 99.8$\pm$0.0 & 99.5$\pm$0.2 & 97.6$\pm$0.4 & \textcolor{violet}{\textbf{100.0}$\pm$0.0} & \textcolor{violet}{\textbf{100.0}$\pm$0.0} \\
              & \textbf{Ours(Prec)} & Post-hoc & No & \textcolor{violet}{\textbf{99.9}$\pm$0.0} & \textcolor{violet}{\textbf{99.5}$\pm$0.2} &97.7$\pm$0.5 &  \textcolor{violet}{\textbf{100.0}$\pm$0.0} & \textcolor{violet}{\textbf{100.0}$\pm$0.0} \\
    \toprule
    \textbf{ID Data}\ \&\ \textbf{Model} & \textbf{Methods} & \textbf{Settings} & \textbf{Additional Data} & \textbf{SVHN}& \textbf{FMNIST} & \textbf{LSUN}& \textbf{TinyImageNet} &\textbf{Corrupted} \\
    \hline
    CIFAR10 & Base Model(Ent) & Traditional & No & 86.4$\pm$4.6 & 90.8$\pm$1.3 & 89.0$\pm$0.5 & 87.5$\pm$1.1 & 85.9$\pm$8.2  \\
    VGG16   & Base Model(MaxP) & Traditional & No & 86.3$\pm$4.4 & 90.4$\pm$1.2 & 88.7$\pm$0.5 & 87.3$\pm$1.1 & 85.7$\pm$8.1   \\
              & MCDropout & Traditional & No & 75.6$\pm$1.1 & 80.9$\pm$0.5 & 85.1$\pm$1.6 & 80.0$\pm$1.3 & 86.8$\pm$1.1   \\
              & BeBayesian &Traditional & No & 91.7$\pm$3.9 & 88.3$\pm$0.6 & 83.2$\pm$0.6 & 81.0$\pm$0.6 & 94.6$\pm$4.6   \\
              & PostNet & Traditional & No & 93.7$\pm$1.2 & 95.8$\pm$2.8 & 93.3$\pm$3.4 & 92.4$\pm$2.9 & 93.4$\pm$2.2  \\
              & ALOE & Traditional & Yes & 99.9$\pm$1.1 & 98.3 $\pm$ 0.7 & 92.1$\pm$1.3  & 99.9$\pm$0.1  & 96.2$\pm$0.7  \\
    \hline
              & \textbf{Ours(Ent)} & Post-hoc & No & 96.3$\pm$3.0 &89.0$\pm$5.2 & 89.6$\pm$3.4 &  89.4$\pm$3.5& 95.9$\pm$4.3 \\
              & \textbf{Ours(MaxP)} & Post-hoc & No & 95.6$\pm$3.6 & 87.8$\pm$4.4 &89.1$\pm$2.4 &  88.2$\pm$2.6 & 94.0$\pm$7.3 \\
              & \textbf{Ours(MI)} & Post-hoc & No & \textcolor{violet}{\textbf{100.0$\pm$0.0}} & \textcolor{violet}{\textbf{98.8$\pm$0.5}} & 95.2$\pm$0.9 &  \textcolor{violet}{\textbf{98.1$\pm$0.3}} & \textcolor{violet}{\textbf{100.0$\pm$0.0}} \\
              & \textbf{Ours(Dent)} & Post-hoc & No & \textcolor{violet}{\textbf{100.0$\pm$0.0}} & 98.4$\pm$0.9 & \textcolor{violet}{\textbf{95.7}$\pm$0.8} &  97.7$\pm$0.5 & \textcolor{violet}{\textbf{100.0$\pm$0.0}} \\
               & \textbf{Ours(Prec)} & Post-hoc & No & \textcolor{violet}{\textbf{100.0$\pm$0.0}} & \textcolor{violet}{\textbf{98.8$\pm$0.5}} & 95.1$\pm$0.5 &  \textcolor{violet}{\textbf{98.1$\pm$0.3}} & \textcolor{violet}{\textbf{100.0$\pm$0.0}} \\
    \toprule
    CIFAR100 & Base Model(Ent) & Traditional & No & 76.2$\pm$5.2 & 77.8$\pm$2.4 & 80.1$\pm$0.5 & 79.7$\pm$0.3 & 65.8$\pm$11.4  \\
    WideResNet   & Base Model(MaxP) & Traditional & No & 73.9$\pm$4.3 & 76.4$\pm$2.3 & 78.7$\pm$0.5 & 78.0$\pm$0.2 & 63.8$\pm$10.4   \\
              & MCDropout & Traditional & No & 77.6$\pm$1.9 & 77.1$\pm$0.4 & 77.0$\pm$3.5 & 79.7$\pm$0.4 & 64.4$\pm$8.9   \\
              & BeBayesian &Traditional & No & 74.9$\pm$8.2 & 81.0$\pm$2.1 & 80.0$\pm$0.7 & 79.6$\pm$0.2 & 63.0$\pm$13.7   \\
              & PostNet & Traditional & No & 83.4$\pm$1.6 & 83.1$\pm$3.2 & 81.0$\pm$1.3 & 80.1$\pm$1.1 & 87.7$\pm$4.2  \\
              & ALOE & Traditional & Yes & 93.9$\pm$0.3 & 86.9$\pm$3.7 & 70.2$\pm$1.2 & 85.3$\pm$3.7 & 92.2$\pm$1.5  \\
    \hline
              & \textbf{Ours(Ent)} & Post-hoc & No & 92.6$\pm$2.0 & 80.8$\pm$3.0 & 81.1$\pm$1.2 &  84.9$\pm$1.2& 85.6$\pm$3.9 \\
              & \textbf{Ours(MaxP)} & Post-hoc & No & 88.6$\pm$3.2 & 78.4$\pm$3.0 & 79.7$\pm$0.6 &  82.4$\pm$0.6& 79.3$\pm$4.5 \\
              & \textbf{Ours} & Post-hoc & No & 94.3$\pm$1.0 & \textcolor{violet}{\textbf{84.4}$\pm$1.8} & 81.9$\pm$4.7 &  85.5$\pm$3.6 & \textcolor{violet}{\textbf{90.8}$\pm$3.3} \\
              & \textbf{Ours(Dent)} & Post-hoc & No & 93.3$\pm$1.4 & 84.0$\pm$2.3 & 79.5$\pm$3.0 &  84.6$\pm$2.8& 89.8$\pm$2.6 \\
              & \textbf{Ours(Prec)} & Post-hoc & No & \textcolor{violet}{\textbf{94.4}$\pm$1.0} & \textcolor{violet}{\textbf{84.4}$\pm$1.7} & \textcolor{violet}{\textbf{82.1}$\pm$4.9} &  \textcolor{violet}{\textbf{85.6}$\pm$3.6}&
              \textcolor{violet}{\textbf{90.8}$\pm$3.4} \\
    \toprule
  \end{tabular}
  \label{table:OOD-appendix}
  \end{center}
  \vspace{-1.5em}
\end{table*}

\begin{table*}[h]
  \begin{center}
  \scriptsize
  \caption{\textbf{OOD Detection AUPR score.}  MI, Dent, and Prec stand for different epistemic uncertainty metrics, i.e., Mutual Information, Differential Entropy, and precision. Settings stand for post-hoc or traditional, i.e., training the entire model from scratch. Additional Data stands for if using additional training data or not.}
  \vspace{-1.5em}
  \begin{tabular}{lllllllll}
    \\
    \toprule
    \textbf{ID Data}\ \&\ \textbf{Model} & \textbf{Methods} & \textbf{Settings} & \textbf{Additional Data} & \textbf{Omniglot}& \textbf{FMNIST} & \textbf{KMNIST}& \textbf{CIFAR10} &\textbf{Corrupted}\\
    \hline
    MNIST & Base Model(Ent) & Traditional & No & 98.7$\pm$0.6 & 97.6$\pm$0.8 & 95.4$\pm$0.7 & 99.4$\pm$0.2 & 99.5$\pm$0.3  \\
    LeNet & Base Model(MaxP) & Traditional & No & 98.4$\pm$0.7 & 97.3$\pm$0.9 & 95.1$\pm$0.8 & 99.2$\pm$0.2 & 99.3$\pm$0.4  \\
              & MCDropout & Traditional & No & 98.1$\pm$0.4 & 97.9$\pm$0.5 & 92.7$\pm$0.8 & 99.3$\pm$0.3 & 98.7$\pm$0.6  \\
              & BeBayesian &Traditional & No & 99.1$\pm$0.4 & 98.4$\pm$0.5 & 95.8$\pm$0.9 & 99.5$\pm$0.2 & 99.6$\pm$0.2  \\
              & Whitebox & Post-hoc & Yes & 97.8$\pm$0.3 & 97.2$\pm$0.6 & 95.3$\pm$0.3 & 99.4$\pm$0.2 & 99.4$\pm$0.2  \\
              & LULA & Post-hoc & Yes & 99.8$\pm$0.0 & 99.5$\pm$0.1 & \textcolor{violet}{\textbf{99.3}$\pm$0.0} & 99.9$\pm$0.0 & 99.9$\pm$0.0  \\
    \hline
              & \textbf{Ours(Ent)} & Post-hoc & No & 99.7$\pm$0.1 & 99.4$\pm$0.2 & 97.9$\pm$0.3 &  \textcolor{violet}{\textbf{100.0}$\pm$0.0} & \textcolor{violet}{\textbf{100.0}$\pm$0.0} \\
              & \textbf{Ours(MaxP)} & Post-hoc & No & 99.2$\pm$0.2 & 99.2$\pm$0.2 &97.6$\pm$0.3 &  \textcolor{violet}{\textbf{100.0}$\pm$0.0} & \textcolor{violet}{\textbf{100.0}$\pm$0.0} \\
              & \textbf{Ours(MI)} & Post-hoc & No & \textcolor{violet}{\textbf{99.9}$\pm$0.0} & \textcolor{violet}{\textbf{99.6}$\pm$0.2} & 97.5$\pm$0.4 & \textcolor{violet}{\textbf{100.0}$\pm$0.0} & \textcolor{violet}{\textbf{100.0}}$\pm$0.0 \\
              & \textbf{Ours(Dent)} & Post-hoc & No & 99.7$\pm$0.0 & 99.4$\pm$0.3 & 97.4$\pm$0.4 &  \textcolor{violet}{\textbf{100.0}$\pm$0.0} & \textcolor{violet}{\textbf{100.0}$\pm$0.0} \\
              & \textbf{Ours(Prec)} & Post-hoc & No & \textcolor{violet}{\textbf{99.9}$\pm$0.0} & \textcolor{violet}{\textbf{99.6}$\pm$0.2} &97.6$\pm$0.5 &  \textcolor{violet}{\textbf{100.0}$\pm$0.0} & \textcolor{violet}{\textbf{100.0}$\pm$0.0} \\
    \toprule
    \textbf{ID Data}\ \&\ \textbf{Model} & \textbf{Methods} & \textbf{Settings} & \textbf{Additional Data} & \textbf{SVHN}& \textbf{FMNIST} & \textbf{LSUN}& \textbf{TinyImageNet} &\textbf{Corrupted} \\
    \hline
    CIFAR10 & Base Model(Ent) & Traditional & No & 77.1$\pm$7.8 & 87.9$\pm$1.5 & 85.2$\pm$0.6 & 83.1$\pm$1.4 & 76.5$\pm$14.0  \\
    VGG16   & Base Model(MaxP) & Traditional & No & 77.4$\pm$7.1 & 86.8$\pm$1.5 & 84.3$\pm$0.5 & 82.5$\pm$1.5 & 76.3$\pm$13.7   \\
              & MCDropout & Traditional & No & 61.5$\pm$1.0 & 70.9$\pm$0.7 & 79.3$\pm$1.9 & 71.9$\pm$1.2 & 78.2$\pm$2.4   \\
              & BeBayesian & Traditional & No & 86.5$\pm$6.6 & 86.9$\pm$0.7 & 82.3$\pm$0.7 & 80.0$\pm$0.5 & 90.8$\pm$8.5   \\
              & Whitebox & Post-hoc & Yes & 92.8$\pm$1.8 & 93.4$\pm$2.4 & 85.3$\pm$2.8 & 86.8$\pm$3.0 & 90.5$\pm$2.1   \\
              & LULA & Post-hoc & Yes & 97.3$\pm$1.4 & 94.8$\pm$0.0 & 93.4$\pm$0.0 & 89.2$\pm$0.0 & 97.7$\pm$1.7   \\
    \hline
              & \textbf{OursELBO(Ent)} & Post-hoc & No & 94.0$\pm$5.7 & 87.0$\pm$5.9 & 87.2$\pm$5.0 &  88.6$\pm$4.6 & 92.3$\pm$8.2  \\
              & \textbf{OursELBO(MaxP)} & Post-hoc & No & 93.6$\pm$5.0 & 86.2$\pm$4.2 & 86.7$\pm$3.5 &  87.5$\pm$3.1 & 90.4$\pm$9.1  \\
              & \textbf{OursELBO} & Post-hoc & No & 100.0$\pm$0.0 & \textcolor{violet}{\textbf{98.4$\pm$0.8}} & 92.9$\pm$1.8 &  \textcolor{violet}{\textbf{97.9$\pm$0.4}} & \textcolor{violet}{\textbf{100.0$\pm$0.0}}  \\
              & \textbf{OursELBO(Dent)} & Post-hoc & No & \textcolor{violet}{\textbf{100.0}$\pm$0.0} & 98.1$\pm$0.9 & \textcolor{violet}{\textbf{94.2}$\pm$1.1} &  97.8$\pm$0.4 & \textcolor{violet}{\textbf{100.0}$\pm$0.0} \\
              & \textbf{OursELBO(Prec)} & Post-hoc & No & 100.0$\pm$0.0 & 98.3$\pm$0.9 & 92.8$\pm$1.9 &  \textcolor{violet}{\textbf{97.9$\pm$0.4}} & \textcolor{violet}{\textbf{100.0$\pm$0.0}}  \\
    \hline
    CIFAR100 & Base Model(Ent) & Traditional & No & 70.8$\pm$5.7 & 73.0$\pm$1.9 & 75.2$\pm$1.1 & 76.6$\pm$0.5 & 60.5$\pm$11.9  \\
    WideResNet   & Base Model(MaxP) & Traditional & No& 67.4$\pm$4.4 & 71.0$\pm$1.9 & 73.3$\pm$0.9 & 73.8$\pm$0.4 & 57.7$\pm$10.1   \\
              & MCDropout & Traditional & No & 67.9$\pm$2.5 & 70.4$\pm$1.5 & 72.4$\pm$3.3 & 75.6$\pm$0.7 & 55.9$\pm$6.3   \\
              & BeBayesian & Traditional & No & 67.5$\pm$7.8 & 75.5$\pm$2.6 & 74.4$\pm$1.0 & 74.9$\pm$0.4 & 58.1$\pm$11.7   \\
              & Whitebox & Post-hoc & Yes & 84.2$\pm$2.0 & 79.5$\pm$0.7 & 74.3$\pm$1.2 & 73.5$\pm$0.5 & 77.1$\pm$4.0   \\
              & LULA & Post-hoc & Yes & 84.3$\pm$1.1 & 83.8$\pm$0.2 & 79.4$\pm$0.4 & 76.8$\pm$0.2 & 79.2$\pm$1.5   \\
    \hline
              & \textbf{OursELBO(Ent)} & Post-hoc & No & 88.0$\pm$3.0 & 75.5$\pm$3.3 & 75.3$\pm$1.7 &  82.0$\pm$1.7& 80.0$\pm$6.3 \\
              & \textbf{OursELBO(MaxP)} & Post-hoc & No & 83.5$\pm$4.6 & 73.7$\pm$2.9 & 74.2$\pm$1.1 &  79.2$\pm$0.9& 71.8$\pm$6.3 \\
              & \textbf{OursELBO(MI)} & Post-hoc & No & 90.1$\pm$1.9 & \textcolor{violet}{\textbf{79.4}$\pm$1.9} & 77.1$\pm$3.6 &  83.2$\pm$4.5 & \textcolor{violet}{\textbf{86.4}$\pm$5.2} \\
              & \textbf{OursELBO(Dent)} & Post-hoc & No & 88.5$\pm$2.2 & 78.6$\pm$2.8 & 74.1$\pm$3.7 &  82.0$\pm$3.6& 85.4$\pm$5.1 \\
              & \textbf{OursELBO(Prec)} & Post-hoc & No & \textcolor{violet}{\textbf{90.2}$\pm$1.9} & \textcolor{violet}{\textbf{79.4}$\pm$1.8} & \textcolor{violet}{\textbf{77.4}$\pm$4.2} &  \textcolor{violet}{\textbf{83.3}$\pm$4.6}& \textcolor{violet}{\textbf{86.4}$\pm$5.2} \\
    \toprule
    
  \end{tabular}
  \label{table: OOD-aupr}
  \end{center}
  \vspace{-1.5em}
 \end{table*}

\subsection{Misclassification Detection}
In the following, we compare our proposed method with several SOTA uncertainty quantification methods with traditional settings on the misclassfication detection task. The results are shown in~Table \ref{table:miss-class-appendix}.

\begin{table*}[t]
  \begin{center}
  \scriptsize
  \captionsetup{font=small}
  \caption{\textbf{Misclassification Results.} Ent and MaxP stand for Entropy and Max Probability, respectively. Settings stand for post-hoc or traditional, i.e., training the entire model from scratch.}
  \vspace{-1.5em}
  \begin{tabular}{llllllllll}
    \\
    \toprule
    \textbf{Methods} & \textbf{Settings} & \textbf{Metric} & \textbf{MNIST}& \makecell{\textbf{CIFAR 10}} & \makecell{\textbf{CIFAR 100}}& \textbf{Metric} &  \textbf{MNIST}& \makecell{\textbf{CIFAR 10}} & \makecell{\textbf{CIFAR 100}}\\
    \hline
    Base Model(Ent)  & Traditional & AUROC & 96.7$\pm$0.9 & 92.1$\pm$0.2 & 87.2$\pm$0.2 & AUPR & 37.4$\pm$4.0 & 47.5$\pm$1.2 & 67.0$\pm$1.2  \\
    Base Model(MaxP) & Traditional & AUROC & 96.7$\pm$0.9 & 92.1$\pm$0.2 & 86.8$\pm$0.2 & AUPR & 39.4$\pm$3.6 & 46.6$\pm$1.1 & 65.7$\pm$1.0  \\
    MCDropout & Traditional & AUROC & 95.1$\pm$0.1 & 90.4$\pm$0.6 & 83.7$\pm$0.1 & AUPR & 34.0$\pm$0.5 & 42.1$\pm$2.1 & 60.3$\pm$0.2  \\
    BeBayesian & Traditional & AUROC & 97.1$\pm$0.5 & 86.6$\pm$0.5 & 85.1$\pm$0.6 & AUPR & 39.4$\pm$4.0 & 47.1$\pm$1.6 & 63.2$\pm$1.3  \\
    PostNet &  Traditional & AUROC & 97.0$\pm$0.6 & 90.3$\pm$0.3 & 84.1$\pm$0.5 & AUPR & 41.3$\pm$2.7 & 51.3$\pm$2.1 & 66.4$\pm$0.3 \\
    \hline
    \textbf{Ours(Ent)} &Post-hoc & AUROC  & 96.9$\pm$0.6 & 91.1$\pm$0.2 & 83.4$\pm$0.1 & AUPR & 35.6$\pm$4.5 & 50.0$\pm$3.1 & 66.3$\pm$0.4  \\
    \textbf{Ours(MaxP)} &Post-hoc & AUROC  &  \textcolor{violet}{\textbf{97.4$\pm$0.4}} &  \textcolor{violet}{\textbf{92.2$\pm$0.7}} &  \textcolor{violet}{\textbf{85.8$\pm$0.2}} & AUPR & \textcolor{violet}{\textbf{44.5}$\pm$5.1} & \textcolor{violet}{\textbf{54.2}$\pm$3.2} & \textcolor{violet}{\textbf{68.2}$\pm$0.5}  \\
    \toprule
  \end{tabular}
  \label{table:miss-class-appendix}
  \end{center}
  \vspace{-1.5em}
\end{table*}

\subsection{Ablation Study}
In the following, we show the complete results of ablation study in Table~~\ref{table:ablation-structure-appendix}.
\begin{table*}[!t]
  \begin{center}
  \scriptsize
  \captionsetup{font=small}
  \caption{\textbf{Ablation Study of Meta-model (CIFAR10 AUROC score)}}
  \vspace{-1.5em}
  \begin{tabular}{llllllll}
    \\
    \toprule
    \textbf{ID Data}\ \&\ \textbf{Model} & \textbf{Methods} & \textbf{SVHN}& \textbf{FMNIST} & \textbf{LSUN}& \textbf{TIM} &\textbf{Corrupted} \\
    \hline
    CIFAR10 & Base Model(Ent)  & 86.4$\pm$4.6 & 90.8$\pm$1.3 & 89.0$\pm$0.5 & 87.5$\pm$1.1 & 85.9$\pm$8.2  \\
    VGG16   & Base Model(MaxP)  & 86.3$\pm$4.4 & 90.4$\pm$1.2 & 88.7$\pm$0.5 & 87.3$\pm$1.1 & 85.7$\pm$8.1   \\
    \hline
            & \textbf{Single-layer Meta(Ent)}  & 90.4$\pm$0.7 & 91.3$\pm$0.2 & 91.5$\pm$0.2 & 89.5$\pm$0.2 & 90.4$\pm$0.8 \\
            & \textbf{Single-layer Meta(MaxP)}  & 90.1$\pm$0.6 & 91.5$\pm$0.1 & 91.4$\pm$0.2 & 89.6$\pm$0.1 & 90.1$\pm$0.6 \\
            & \textbf{Single-layer Meta(MI)}  & 90.6$\pm$1.2 & 90.1$\pm$0.6 & 91.2$\pm$0.2 & 88.8$\pm$0.3 & 90.5$\pm$1.3 \\
            & \textbf{Single-layer Meta(Dent)}  & 90.4$\pm$0.9 & 90.7$\pm$0.3 & 91.4$\pm$0.2 & 89.2$\pm$0.2 & 90.3$\pm$1.0 \\
            & \textbf{Single-layer Meta(Prec)} & 90.6$\pm$1.2 & 90.0$\pm$0.6 & 91.2$\pm$0.2 & 88.8$\pm$0.4 & 90.6$\pm$1.4 \\
    \hline
      & \textbf{Cross-Ent Meta(Ent)}  & 94.2$\pm$2.2 & 91.2$\pm$0.7 & 91.2$\pm$0.6 &  90.3$\pm$0.8& 94.7$\pm$2.5 \\
        & \textbf{Cross-Ent Meta(MaxP)}  & 93.3$\pm$1.6 & 91.1$\pm$0.7 &90.9$\pm$0.4 &  90.0$\pm$0.6 & 94.0$\pm$2.0 \\
    \hline
       & \textbf{Base-Model+LastLayer(Ent)}& 93.0$\pm$1.1 & 90.2$\pm$0.3 & 91.9$\pm$0.2 &  89.9$\pm$0.2& 93.1$\pm$1.2 \\
        &\textbf{Base-Model+LastLayer(MaxP)} & 92.9$\pm$0.9 & 90.5$\pm$0.3 &91.9$\pm$0.2 &  90.1$\pm$0.1 & 93.1$\pm$0.9 \\
    \hline        
    & \textbf{$10\%$data(Ent)}  & 90.0$\pm$7.1 &89.1$\pm$4.4 & 88.7$\pm$6.4 &  88.2$\pm$5.2& 90.2$\pm$7.4 \\
              & \textbf{$10\%$data(MaxP)} & 90.9$\pm$5.5 & 88.1$\pm$3.3 &86.9$\pm$7.0 &  86.5$\pm$4.5 & 91.7$\pm$5.4 \\
              & \textbf{$10\%$data(MI)}& 99.9$\pm$0.2 & 98.1$\pm$2.2 & 95.4$\pm$1.7 &  97.2$\pm$2.4 & 99.9$\pm$0.1 \\
              & \textbf{$10\%$data(Dent)}  & 96.7$\pm$3.5 & 97.4$\pm$3.0 & 94.3$\pm$4.1 &  95.7$\pm$4.1 & 96.7$\pm$3.4 \\
               & \textbf{$10\%$data(Prec)}  & 99.9$\pm$0.1 & 98.0$\pm$2.2 & 95.4$\pm$1.7 &  97.3$\pm$2.3 & 99.9$\pm$0.1 \\
    \hline
              & \textbf{Ours(Ent)} & 96.3$\pm$3.0 &89.0$\pm$5.2 & 89.6$\pm$3.4 &  89.4$\pm$3.5& 95.9$\pm$4.3 \\
              & \textbf{Ours(MaxP)}  & 95.6$\pm$3.6 & 87.8$\pm$4.4 &89.1$\pm$2.4 &  88.2$\pm$2.6 & 94.0$\pm$7.3 \\
              & \textbf{Ours(MI)}  & \textcolor{violet}{\textbf{100.0$\pm$0.0}} & \textcolor{violet}{\textbf{98.8$\pm$0.5}} & 95.2$\pm$0.9 &  \textcolor{violet}{\textbf{98.1$\pm$0.3}} & \textcolor{violet}{\textbf{100.0$\pm$0.0}} \\
              & \textbf{Ours(Dent)}  & \textcolor{violet}{\textbf{100.0$\pm$0.0}} & 98.4$\pm$0.9 & \textcolor{violet}{\textbf{95.7}$\pm$0.8} &  97.7$\pm$0.5 & \textcolor{violet}{\textbf{100.0$\pm$0.0}} \\
               & \textbf{Ours(Prec)} & \textcolor{violet}{\textbf{100.0$\pm$0.0}} & \textcolor{violet}{\textbf{98.8$\pm$0.5}} & 95.1$\pm$0.5 &  \textcolor{violet}{\textbf{98.1$\pm$0.3}} & \textcolor{violet}{\textbf{100.0$\pm$0.0}} \\
    \toprule
  \end{tabular}
  \label{table:ablation-structure-appendix}
  \end{center}
  \vspace{-1.5em}
\end{table*}

\end{document}